 \documentclass[final,5p,times,twocolumn]{elsarticle}

\usepackage{amssymb}
\usepackage{amsmath}

\usepackage[ruled]{algorithm2e}                 
\usepackage{amssymb}
\usepackage{tabularx}
\usepackage{color}
\usepackage{multicol}
\usepackage{multirow}
\usepackage{makecell}
\usepackage{longtable}
\usepackage{booktabs}

\journal{Knowledge-Based Systems}

\begin{document}

\begin{frontmatter}



\title{Instructing the Architecture Search for Spatial-temporal Sequence Forecasting with LLM
}


\author[1,2]{Xin Xue}
\ead{xuexin@act.buaa.edu.cn}
\author[2,3]{Haoyi Zhou} 
\ead{haoyi@buaa.edu.cn}
\author[1,2]{Tianyu Chen}
\ead{tianyuc@buaa.edu.cn}
\author[4]{Shuai Zhang}
\ead{zhangs@act.buaa.edu.cn}
\author[1,2]{Yizhou Long}
\ead{longyizhou2001@buaa.edu.cn}
\author[1,2]{Jianxin Li\corref{cor1}} 
\ead{lijx@buaa.edu.cn}
\cortext[cor1]{Corresponding author}

\affiliation[1]{organization={Department of Computer Science, Beihang University},
            city={Beijing},
            postcode={100191}, 
            country={China}}
\affiliation[2]{organization={Beijing Advanced Innovation Center for Big Data and Brain Computing},
            city={Beijing},
            postcode={100191}, 
            country={China}}
\affiliation[3]{organization={Department of Software, Beihang University},
            city={Beijing},
            postcode={100191}, 
            country={China}}
\affiliation[4]{organization={Zhongguancun Laboratory},
            city={Beijing},
            postcode={100190}, 
            country={China}}

\begin{abstract}

Spatial-temporal sequence forecasting (STSF) is a long-standing research problem with widespread real-world applications. Neural architecture search (NAS), which automates the neural network design, has been shown effective in tackling the STSF problem. However, the existing NAS methods for STSF focus on generating architectures in a time-consuming data-driven fashion, which heavily limits their ability to use background knowledge and explore the complicated search trajectory. Large language models (LLMs) have shown remarkable ability in decision-making with comprehensive internal world knowledge, but how it could benefit NAS for STSF remains unexplored. In this paper, we propose a novel NAS method for STSF based on LLM. Instead of directly generate architectures with LLM, We inspire the LLM's capability with a multi-level enhancement mechanism. Specifically, on the step-level, we decompose the generation task into decision steps with powerful prompt engineering and inspire LLM to serve as instructor for architecture search based on its internal knowledge. On the instance-level, we utilize a one-step tuning framework to quickly evaluate the architecture instance and a memory bank to cumulate knowledge to improve LLM's search ability. On the task-level, we propose a two-stage architecture search, balancing the exploration stage and optimization stage, to reduce the possibility of being trapped in local optima. Extensive experimental results demonstrate that our method can achieve competitive effectiveness with superior efficiency against existing NAS methods for STSF.

\end{abstract}


\end{frontmatter}

\section{Introduction}
Spatial-temporal sequence forecasting (STSF) is a long-standing problem in which the cross relations between spatial features and temporal features are complex. 
A wealth of manually designed STSF predictors~\cite{STFGNN,STGCN,GraphWaveNet,GMAN} have been constructed to extract temporal and spatial information in different combination which is related to the specific STSF tasks. However, how to automatically and adaptively determine the combination of the temporal and spatial feature extraction modules according to specific tasks remains an important problem, as various STSF tasks have diverse modeling preferences with mixed spatial-temporal dependencies.
Neural architecture search (NAS) methods~\cite{DARTS,PT,AutoST,AutoCTS+} are designed to find proper neural architectures for specific STSF tasks. The intrinsic problem lies in NAS methods is to balance the time-consumption during the searching process and the expected performance of the searched architecture. The heuristic searching methods~\cite{NAS,NASNet,PNAS} are time-consuming to search over a discrete set of candidate architectures. Avoiding large-scale searching in almost the whole search space, the gradient-based methods~\cite{DARTS,PT,SDarts,PC-darts} relax the searching space to be continuous for efficiency and introducing the learning-based searching mechanism, but it still requires heavy work in policy learning and parameter tuning. Owing to the substantial number of model parameters and the complexity of spatial-temporal correlations, \textbf{how to efficiently find promising architectures for specific STSF tasks is still an important issue.}

\begin{figure*}[t]
  \centering
   \includegraphics[width=\linewidth]{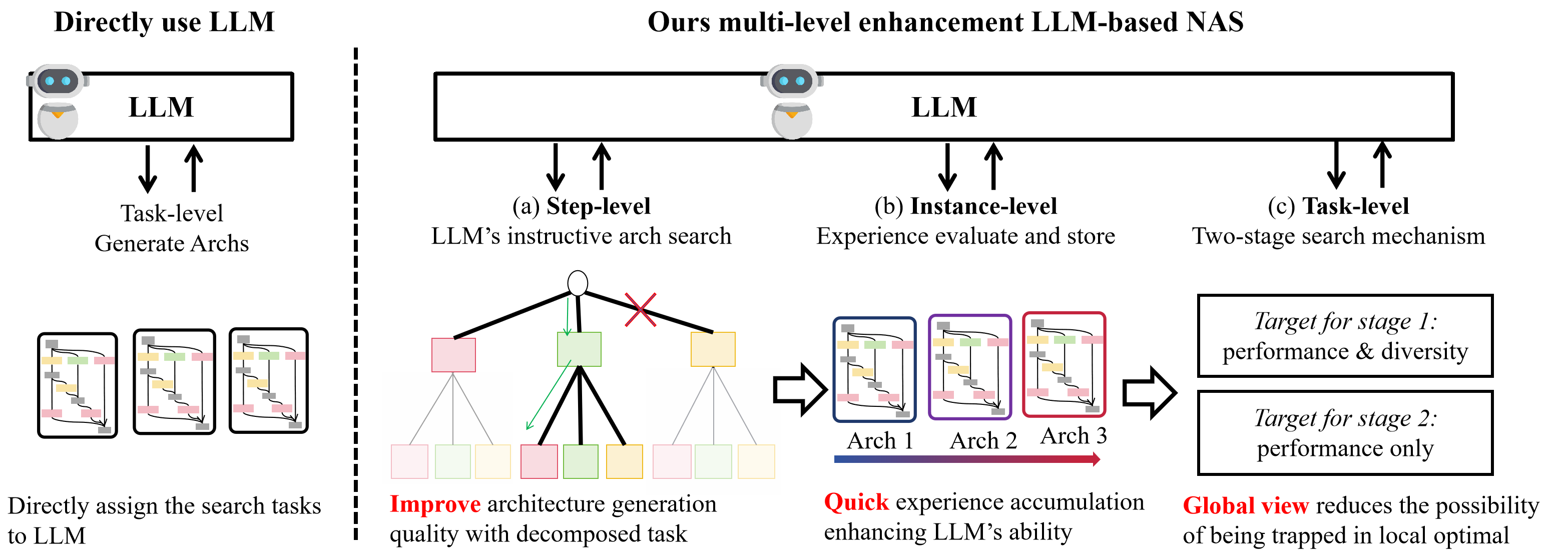}
  \caption{\textbf{Directly use LLM for NAS V.S. Our multi-level LLM-based NAS.} Existing methods simply use the LLM as an end-to-end architecture generator. However, our proposed method involves LLM deeply by multiple-level enhancement to improve and inspire LLM's decision-making ability. (a) Step-level: Decompose the task into logically coherent steps with powerful prompting mechanism to lead LLM to instruct the generation and selection of the candidate architecture based on its internal world knowledge, allowing LLM to quickly find well-performed architectures. (b) Instance-level: A one-step evaluation framework simulates the architecture candidates quickly and accumulate experience with a memory bank to inspire LLM to explore better architectures. (c) Task-level: Two-stage search mechanism assists the large model in determining the optimal results and reduce the possibility of being trapped into local optima.}
  \label{title}
\end{figure*}

Existing NAS methods for STSF show limited ability to use background knowledge and explore the complicated search trajectory, where LLM exactly to be proficient in. Recently, large language models (LLMs) have garnered significant acclaim for their remarkable performance in decision-making tasks, owing to their exceptional comprehension and analytical prowess, even without training on specific tasks. Some LLM-based NAS methods~\cite{AmoebaNet,EvoPrompting,EvolvingLLM,GPT-NAS,GNAS} simplified the search process, thereby enabling the expeditious designing and exploring of diverse architectures. Despite these advancements, the existing methods overlook the immense analytical potential of LLMs, restricting their role to random code generators devoid of the understanding of underlying structural rules. \textbf{Consequently, there remains an untapped opportunity to harness more capabilities of LLMs to enhance the process of neural architecture search for STSF.}

In order to improve neural architecture search for spatial-temporal sequence forecasting (STSF), we propose an LLM-based framework named \underline{I}nstructive neural architecture search for \underline{S}patial \underline{T}emporal \underline{S}equence  forecasting (\textbf{ISTS}), aiming to take advantage of the analytical capabilities of LLM with the enhancement at three different levels, From local to global, we enhance the searching process at the step-level, instance-level and task-level, which is as illustrated in Figure ~\ref{title}. 

At the \textbf{step-level}, we decompose the task into steps by constructing powerful prompts for LLM to make the architecture candidate generation with its internal knowledge, which can enhance LLM's ability in architecture candidate generation and allow the LLM to find well-performed architectures with a small number of searches. At the \textbf{instance-level}, the candidate architecture is simulated and preliminarily evaluated, and the feedback with search experience from an ordered memory bank is provided as the evidence for LLM's decision-making, which benefits LLM to make reasonable decisions with cumulated experience. At the \textbf{task-level}, we propose the two-stage task-wise search mechanism, enabling large models to balance exploration process and optimization process, which reduce the possibility of being trapped by local optimal results. The multi-level enhancement inspire LLM to make sensible decisions on NAS for specific STSF tasks. 

We extensively evaluate our method on real-world datasets. The framework of ISTS can be flexibly adapted to well-known open-source pre-trained large language models. The results indicate that our proposed ISTS methods achieve better or competitive results comparing with both state-of-the-art NAS and non-NAS methods in five STSF datasets. Time-consumption evaluation indicates that our inferring-only LLM-based NAS mechanism shows significant efficient advantages over traditional NAS methods. Additionally, we provide ablation experiments and visualization of the architectures discovered by ISTS, which vividly illustrates its extensive adaptability to various prediction tasks.

Our contributions can be summarized as follows:
\begin{itemize}
    \item We firstly propose a novel LLM-based NAS method \textbf{ISTS}, an automated neural architecture search method instructed by LLM for specific STSF tasks, which can significantly exploit the potential of large language models to be acknowledged from background knowledge and discover hidden policy.
    
    \item We dives into the question of how to comprehensively enhance LLM's comprehension ability for architecture searching more than directly asking LLM, by designing a multi-level enhancement, i.e. step-level, task-level, and instance-level. The enhancements provide a chance of  decomposing the architecture generation task into logically coherent steps and quickly simulating candidates, which can improve the generation quality and efficiency. 
    
    \item ISTS provides a better balance of efficiency and performance. Experimental results demonstrate that ISTS can achieve competing or better searching results for STSF architectures with significantly reduction of the processing time.

\end{itemize}
\section{Related Work}

\subsection{Spatial-temporal Sequence forecasting} Methods for spatial-temporal sequence forecasting typically combine spatial operators and temporal operators in different ways to capture various spatial-temporal dependencies. Spatial operators usually consist of GCN-based methods~\cite{DGCN-LSTM-ST,AutoST,DGCN-RNN}, while temporal operators usually consist of RNN-based methods~\cite{DGCN-LSTM-ST,DGCN-RNN} and attention-based methods~\cite{AutoST}. These operators can be combined in different orders, namely spatial-first~\cite{STGCN}, temporal-first~\cite{GraphWaveNet}, and spatial-temporal synchronous~\cite{STSGCN}, which vary for specific tasks. However, manually designed architectures face difficulty in accurately adapting various spatial-temporal dependencies in different STSF tasks. 


\subsection{Traditional Neural Architecture Search} 

Traditional methods can be categorized into heuristic searching methods~\cite{NAS,NASNet,PNAS} and gradient-based methods~\cite{ENAS,DARTS,PT,SGAS}. 

NAS~\cite{NAS} is the pioneer exploration which solves this problem with recurrent neural networks trained to perform parameter predicting. Instead of searching separated parameters, NASNet~\cite{NASNet} introduces the concept of searching cells to balance efficiency and performance. Tired of large-scale exhaustive search, PNAS~\cite{PNAS} uses a predictor to choose promising candidates. AmoebaNet~\cite{AmoebaNet} uses an evolution method to explore and generate architectures, instead of an exhaustive search. To further reduce the search space, ENAS~\cite{ENAS} converts the searching tree into a directed acyclic graph (DAG) with operation cells. 

DARTS~\cite{DARTS} utilizes the thoughts of the dual graph to reduce the searching space and relax the discrete searching set to be continuous, which firstly enables gradient-based searching. Gradient-based methods significantly reduce time-consumption and hardware-demand with superiority performance. Optimizations for gradient-based methods~\cite{SGAS,PT}  can improve the performance of gradient-based neural architecture search. AutoST~\cite{AutoST} proposed an automatic search strategy for spatio-temporal forecasting by determining the mixed spatial-temporal structure. Although these works tried to reduce the search space, the policy learning is still heavy work. 

\subsection{LLM-based Neural Architecture Search} 

With the powerful generation capability, LLM-based methods have appeared in recent years. EvoPrompting~\cite{EvoPrompting} uses a large language model as a code generator in architecture evolution. GPT-NAS~\cite{GPT-NAS} reform the LLM-based code generator with evolutionary thought, by eliminating a specific layer and asking LLM to rebuild the model. GNAS~\cite{GNAS} utilizes the power of LLMs to accelerate the architecture generation process in graph tasks. However, existing LLM-based searching methods treat LLMs as random code generators without prompting skills and mechanisms for improvement, therefore cannot fully unlock the potentials of LLM in NAS. Besides, how LLM-based NAS can be utilized for spatial-temporal forecasting remains unexplored in the literature.

\section{Problem Definition}
\subsection{Spatial-temporal Sequence Forecasting}
Spatial-temporal sequence forecasting (STSF) is to forecast a set of temporal sequences with spatial correlations. The spatial correlation is often shown in a graph $G=(V, E)$, where $V$ and $E$ are the node sets and edge sets, and $N$ is the size of the node set. $\mathbf{X}=\{\mathbf{X}_1, \mathbf{X}_2, ...,\mathbf{X}_T\}$ is a set of temporal sequences of $T$ time steps, where $X \in \mathbb{R}^{T\times N \times C}$, and $C$ is the size of feature dimension. $\mathbf{X}_t$ is a vector containing the observing values of all the nodes $v \in V$ of the timestamp $t$. The STSF task is to find a learning function $f$ for spatial-temporal sequence prediction, and $f$ is formulated as follows:
\begin{equation}
    f:[\mathbf{X}_{(t-s):t}, G] \stackrel{f}{\longrightarrow}\mathbf{X}_{(t+1):(t+P)},
\end{equation}
where $\mathbf{X}_{(t-s):t}$ is the history observation with length S, and $\mathbf{X}_{(t+1):(t+P)}$ is the future sequence with length $P$.

\subsection{Neural Architecture Search for STSF}

Neural Architecture Search (NAS) for STSF is to generate neural architectures $A$ to formulate the learning function $f$ automatically. The problem can be translated to a bi-level optimization problem as follows:
\begin{equation}
    \begin{aligned}
        \min_{A} &\nabla \mathcal{L} _{val}(w^{*}(A), A) \\
        s.t. &\ w^{*}(A) = \mathop{\arg\min}\limits_{w} \mathcal{L}_{train}(w(A), A)
    \end{aligned} ~,
\end{equation}
where $\mathcal{L} _{val}$ and $\mathcal{L}_{train}$ denotes the validation and training loss, and $w$ denotes weights of the searched architecture $A$.  

In this paper, the goal is to discover neural architectures tailored to the specific problem of STSF efficiently and effectively.

\section{Method}

We present the overview of the framework in Figure ~\ref{architecture}, which illustrates the multi-level enhanced neural architecture search for STSF instructed by LLM. In this section, we first introduce the search space, then elaborate on the three key components of our method. 

\subsection{Search Space}
\label{ch-search-space}

\subsubsection{Cells and Operators}

\label{app-defin}
\begin{figure*}[t]
  \centering
   \includegraphics[width=\linewidth]{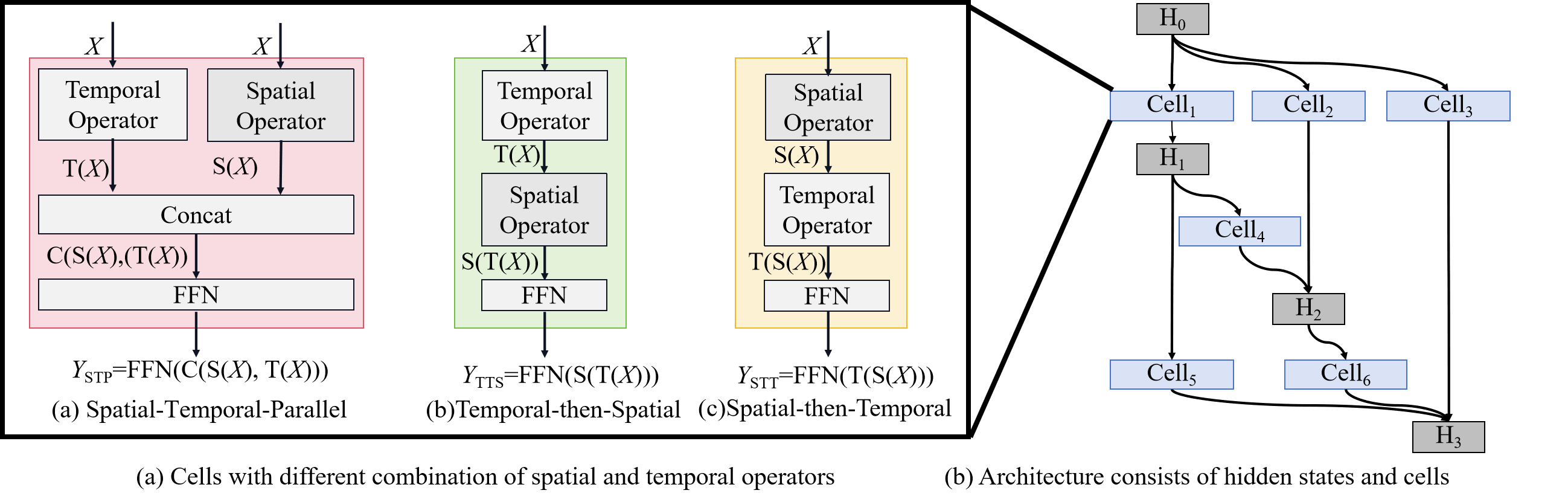}
  \caption{\textbf{Cells and the architecture for searching.} (a) 3 types of cells with different combination of spatial and temporal operators provide possibility to capture various spatial-temporal mixed dependencies. (b) Architecture combines multiple cells in a DAG way, and each cell has 3 kinds of option as indicated in (a). }
  \label{prblem}
\end{figure*}

To adaptively capture various temporal-spatial dependencies in STSF, we incorporate spatial operator and temporal operator in different ways. Specifically, we utilize spatial-temporal-parallel (STP) cell, spatial-then-temporal (STT) cell, and temporal-then-spatial (TTS) cell. The definitions of these cells are as follows:
\begin{equation}
    \begin{aligned}
    Y_{\mathrm{STP}} &= \mathrm{FFN}\big(\mathrm{concat}[\mathrm{S}(X), \mathrm{T}(X)]\big),\\
     Y_{\mathrm{STT}} &= \mathrm{FFN}\big(\mathrm{T}(\mathrm{S}(X))\big),\\
     Y_{\mathrm{TTS}} &= \mathrm{FFN}\big(\mathrm{S}(\mathrm{T}(X))\big),
    \end{aligned}
\end{equation}
where $\mathrm{S}(\cdot)$ indicates spatial operator, $\mathrm{T}(\cdot)$ indicates temporal operator, $\mathrm{FFN}(\cdot)$ indicates feed forward layers. 

We select a GCN-based spatial operator, namely high-order mix graph convolution~\cite{AutoST} to extract spatial features, formulated as Equation~\ref{eqs-MG}, which is powerful in dealing with high-order structure relations. 
\begin{equation}
\label{eqs-MG}
   \mathrm{S}(X, A, ord)  
    = \begin{cases}
	X, & ord = 0\\
	\text{MixGC}(X, A), & ord = 1 \\
        \text{MixGC}(S(X, A, ord - 1), A) & ord > 1
		   \end{cases},
\end{equation}
where 
\begin{equation}
\begin{aligned}
\text{MixGC}(X, A) = \text{ChebNet}(X, A) + \text{AdapDC}(X, A) \\
 = \hat{A}XW_g + P_fXW_f + P_bXW_b + \hat{A}_{adp}XW_{adp},
\end{aligned}
\end{equation}
which are proposed in ChebNet~\cite{ChebNet} and Adaptive Diffusion Convolution~\cite{adapDC}. The spatial connections of nodes are described by the normalized adjacency matrix $\hat{A}$ and a learnable correlation matrix $\hat{A}_{adp}$. $ord$ can be adjusted through hyper-parameters.

We use the technique of POLLA~\cite{Pola}, formulated as Equation~\ref{eq-polla}, which can efficiently capture temporal features with linear complexity, as the temporal operator. 
\begin{equation}
\label{eq-polla}
    T(X) = \text{LA}(\phi(Q), \phi(K), V),
\end{equation}
where $Q = XW_Q, K = XW_K, V = XW_V$, $W_Q$, $W_K$, $W_V$ are learnable matrices, and $\phi(\cdot)$ is a mapping function. 

\subsubsection{Architecture}

\begin{figure*}
  \centering
   \includegraphics[width=\linewidth]{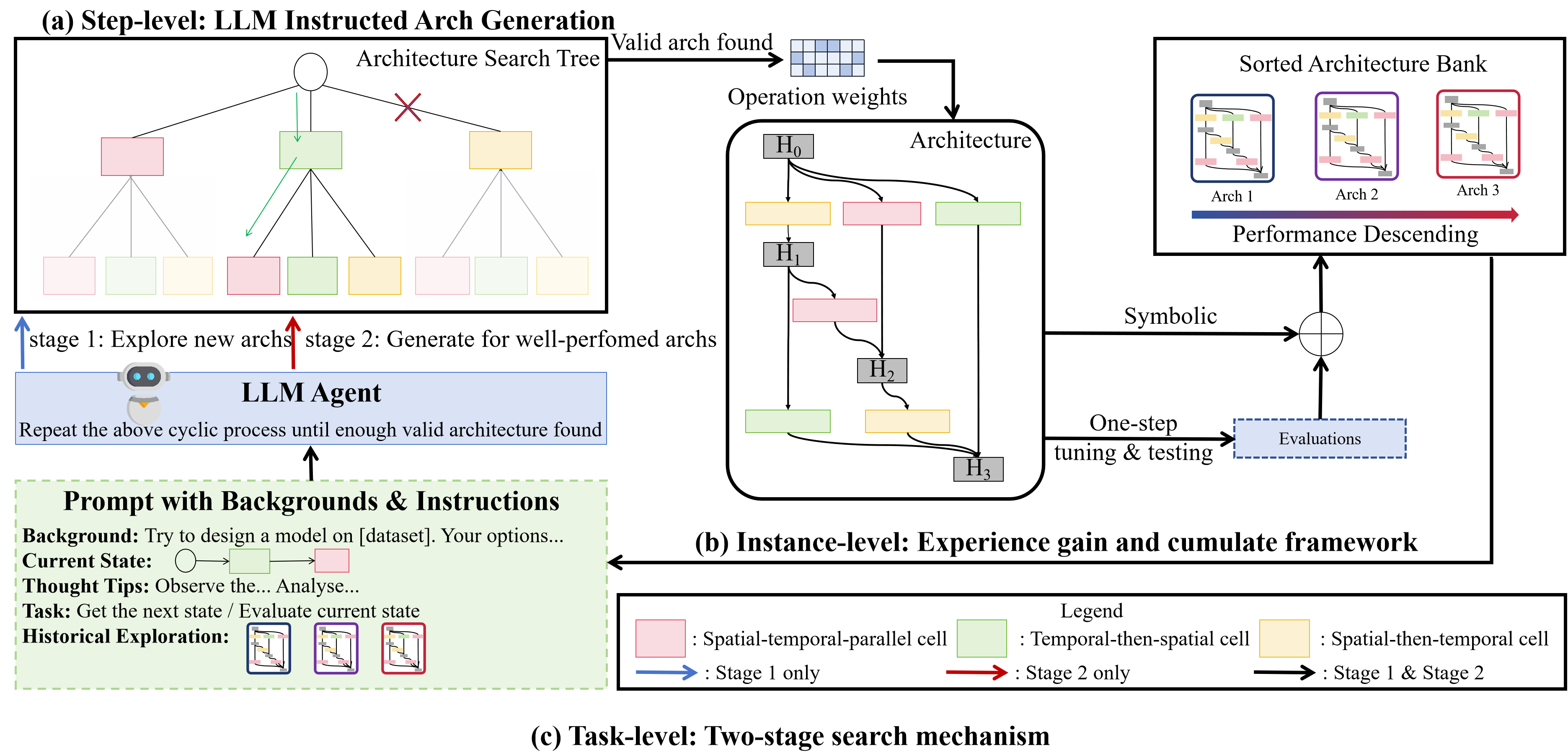}
  \caption{The framework of our proposed ISTS method. (a) LLM instructed architecture generation: Powerful task decomposing prompt encourage LLM to generate new architecture candidates sensibly its policy with LLM's internal knowledge and comprehension and quickly find well-performed architectures. (b) Experience gain and cumulate framework: Once a valid architecture candidate generated, the framework construct the architecture and quickly tune and evaluate it with several training data. The sorted memory bank store the architecture evaluation results to enhance LLM by providing experience accumulation. (c) Two-stage search mechanism: To escape from local optimal results, we encourage LLM to explore new architectures though may not be well-performed ones in the first stage. Then, with a wide exploration, the only target for the second stage is to improve performance. }
  \label{architecture}
\end{figure*}

Figure ~\ref{architecture} shows the structure of neural architecture $A$, with 4 nodes and 6 edges combining as a directed acyclic graph (DAG), where nodes denote hidden states and edges denote alternative cells. The DAG structure provides a flexible way to capture various spatial-temporal dependencies, where jumping and merging actions enable formulating more information. Three types of cells defined as Section~\ref{ch-search-space} with different types of combinations are alternative options for each cell. 



\subsection{LLM Instructed Architecture Generation}

It's convinced that powerful prompting methods with suitable thought decomposition contributes significantly to LLM's comprehensive and decision-ymaking abilities~\cite{COT,TOT}. We employ two prompting mechanism decomposing the generation task into different scale of steps for STSF architecture search. 
\begin{algorithm}


    \caption{LLM Instructed Architecture Generation}
    \label{al-generation}
    \KwIn{COT-style prompts template $P_C$, TOT-style prompts template $P_T=\{P_{T-Gen}  P_{T-Eval}\}$,  Experience from memory bank $E$, Stage strategy $S$}
    \KwOut{Generated architecture $A$.} 

    \eIf{COT-style search} {
         $A \gets LLM(P_C, E, S) $; \\
    } {
         $A\gets \emptyset$; \\
        \While{$A$ is not a valid architecture}{
            New cell $c \gets LLM(P_{T-Gen}, E, S)$; \\
            Get evaluation $e \gets LLM(P_{T-Eval}, E, S)$; \\
            \If{e}{
                $A\gets A + c $; \\
            }
        }
    }

    \Return $A$

\end{algorithm}
\subsubsection{Generation Mechanism}

 We employ two popular and simple prompting mechanism, namely chain-of-thought~\cite{COT} and tree-of-thought~\cite{TOT}, providing different way of task decomposition to assist LLM in understanding the tasks and deal with them step-by-step.

 \textbf{COT-style Architecture Search.} We utilize the chain-of-thought mechanism by decomposing the architecture generation into three steps, including background comprehension, experience observation and candidate generation. Background comprehension is to provide knowledge about downstream tasks, metrics and architecture templates, inspiring large models to extract useful information from background and internal knowledge. Experience observation encourage LLM to observe cumulated experience from the memory bank (details of memory bank in ~\ref{ch-bank}). Finally, we guide LLM to generate new candidates with strategy of current stage (details of strategy in ~\ref{ch-strategy}).

 \textbf{TOT-style Architecture Search.} TOT-style architecture search is to further decompose the intermediate candidate generation process based on COT. Specifically, for the step of candidate generation mentioned in COT-style, instead of generating the whole architecture in one step, TOT-style search generates and evaluates each cell of the architecture in the search tree. We adopt DFS search algorithm for tree search, and guide LLM to generate new cells of the architecture and evaluate the behaviors by itself for several rounds, until a valid architecture is found in the search tree.

The entire process of architecture generation with two kinds of prompt mechanism is outlined in Algorithm ~\ref{al-generation}.

\subsubsection{Prompt Template}

\begin{table}

  \centering
  \begin{tabularx}{\linewidth}{l|X}
    \toprule
    \textbf{Component} & \textbf{Content} \\
    \toprule

Background     &  Dataset description: \textcolor{blue}{[Target Dataset]} \newline Metrics: \textcolor{blue}{[Metric 1]}, \textcolor{blue}{[Metric 2]}... \newline ST-Cells: \textcolor{blue}{[Option 1]}, \textcolor{blue}{[Option 2]}... \newline Response format: \textcolor{blue}{[Format Example]} \\

    \midrule

Instruction & 1. Background Comprehension  \newline 2. Experience Observation \newline 3. Current Task: Generate / Evaluate \\

    \midrule

Experience  & Experience from memory bank \textcolor{blue}{$E$} \\
    \midrule
Strategy  & Strategy of current stage \textcolor{blue}{$S$} \\

    \bottomrule
  \end{tabularx}
    \caption{Outline of the designed prompts. 
    }
  \label{prompt-summary-table}
\end{table}

For background comprehension part, we concentrate background information together to form templates, including introducing the dataset, calculating metrics, defining ST-Cells, and formatting responses. Both the COT-style and TOT-style templates consist of three-step instructions. The difference lies in that the third step of the COT-style template is an instruction, whereas that of the TOT-style template could be generation or evaluation. Experience $E$ and strategy $S$ could be obtained from experience accumulative framework (Section~\ref{ch-bank}) and two-stage mechanism (Section~\ref{ch-strategy}). The outline of the prompt template is shown in Table ~\ref{prompt-summary-table}, and details of the prompts are presented in ~\ref{app-prompts}. 

Guided by powerful prompts, the LLM actively and effectively generate architectures with its internal knowledge and comprehension. This not only reduces the number of search times but also improves the quality of candidates, leading to enhanced effectiveness and efficiency.

\subsection{Experience Accumulative Framework}
\label{ch-bank}

Once a valid candidate architecture $A$ is generated, we construct and simulate the architecture to preliminarily evaluate the architecture and record it as experience to cumulate and enhance the LLM's ability.  

\subsubsection{One-step Tuning and Evaluating}

For a candidate architecture $A$, we construct a combination DAG structure (mentioned in Section~\ref{ch-search-space}) with a set of learnable parameters $w(A)$. The parameters $w(A)$ are trained with a set of training data targeting for the lowest training loss $\mathcal{L}_{train}$, which is formulated as Equation~\ref{eq-train}.

\begin{equation}
    \label{eq-train}
    w^{*}(A) = \mathop{\arg\min}\limits_{w} \mathcal{L}_{train}(w(A), A)
\end{equation}

Then the performance is evaluated by another set of validation data as $\mathcal{L}_{valid}(w^*(A), A)$, where $\mathcal{L}_{valid}$ is a metric depending on downstream tasks. The quick evaluation provides a preliminarily feedback for LLM, enabling it to adjust policy and generate more promising architectures with efficiency ensuring.

\subsubsection{Sorted Architecture Bank}

The architecture bank stores the experience of previous trails. Specifically, the storing set $E$ is initialed with an empty set, and when an evaluation of a candidate is performed, $E$ is appended as Equation ~\ref{eq-bank}.

\begin{equation}
    \label{eq-bank}
    E = E \cup \{(A, \mathcal{L}_{valid}(w^*(A), A))\}
\end{equation}

To avoid LLM to make numerical calculation, which may introduce new errors, and further simplify the problem, we sort the samples in $E$ with descending order of $\mathcal{L}_{valid}$. $E$ is sorted each time after new sample inserted. The cumulative mechanism reserve existing exploring experience, providing accumulated knowledge for LLM to enhance decision-making ability.

\subsection{Two-stage Search Mechanism}
\label{ch-strategy}

To reduce the possibility of being trapped with local optima, we introduce a two-stage search mechanism, which is inspired by the strategy of GNAS~\cite{GNAS}. However, the search strategy in GNAS is guiding LLM as a single part of prompt, and the decision to explore or optimize is made by LLM, lacking outer binding force and flexibility. In our LLM-assisted NAS for STSF, we divide the whole process into two stages, namely exploring stage and optimizing stage. The two stages have different searching target to guide LLM to explore for more promising search space.
\begin{itemize}
    \item \textbf{Exploring Stage.} The first several rounds are set as exploring stage, and the target is to explore more unexplored promising candidates with target prompt $S_1$. 
    \item \textbf{Optimizing Stage.} The following rounds are set as optimizing stage, and the target is to find the candidate architecture with potentially best performance with target prompt $S_2$.
\end{itemize}
Details of the prompt $S_1$ and $S_2$ are presented in ~\ref{app-prompts}.

With two-stage strategy, the procedure of the entire process is outlined in Algorithm ~\ref{al-whole}, conducting architecture search for STSF in the search space defined in ~\ref{ch-search-space} with LLM in a effective and efficient way. 

\begin{algorithm}
\caption{LLM-assisted NAS for STSF}
\label{al-whole}
\KwIn{Prompts Template $P=\{P_C, P_T\}$, Maximum search round $T$, rate for exploration rounds $r$.} 
\KwOut{Searched architecture $A^*$.}

 $\textrm{Initialize memory bank}\ E = \emptyset $; \\

 $t \gets 0$; \\
  
  \While{$t < T$}{
    $t \gets t+1$; \\
    \eIf{Exploring Stage}{
         $A \gets \textrm{LLM\_Instructed\_Arch\_Gen}(P, E, S_1)$; \\
    }{
        $A \gets \textrm{LLM\_Instructed\_Arch\_Gen}(P, E, S_2)$; \\
    }

    $w^{*}(A) \gets \mathop{\arg\min}\limits_{w}\mathcal{L}_{train}(w(A), A)$; \\

    $\mathcal{L}(A) \gets \mathcal{L}_{valid}(w^*(A), A)$; \\
   
    $E \gets E \cup \{(A, \mathcal{L}(A))\}$; \\
    
  }
    $A^*\gets \mathop{\arg\min}\limits_{A}\mathcal{L}(A)$;
    
    \Return $A^*$

\end{algorithm}

\begin{table*}[htbp!]

  \centering
\resizebox{\linewidth}{!}{
  \begin{tabular}{cc|cc|cccc|ccc}
    \toprule
     \multicolumn{2}{c|}{Category} &  \multicolumn{2}{c|}{Non-NAS} & \multicolumn{6}{c}{NAS} \\
     \cmidrule(r){1-2}
    \cmidrule(r){3-4}
    \cmidrule(r){5-10}
    \multicolumn{2}{c|}{Method}  & STSGCN & GraphW.N. & DARTS & DARTS-PT & SDARTS & PC-DARTS & ISTS-COT & ISTS-TOT \\
    \midrule
    \multirow{3}{*}{PEMS-03} & MAE & 18.73 & 15.46 & 15.58  & \textbf{15.36}  & \underline{15.41}  & 15.55  & \underline{15.41}  & \textbf{15.36}   \\ 
    & MAPE &     20.31\% & \textbf{15.41\%} & 16.93\%  & 18.40\%  & 16.57\%  & 16.90\%  & 17.57\%  & 18.27\%   \\ 
    & RMSE &    29.19 & \textbf{24.34} & 26.04  & 26.09  & 25.85  & 26.24  & 26.23  & 26.09   \\ 
    \midrule
    \multirow{3}{*}{PEMS-04} & MAE &    22.11 & 20.74 & 19.86  & 19.31  & 19.40  & 19.64  & \textbf{19.22}  & \underline{19.26}   \\ 
    & MAPE &    16.97\% & 16.56\% & 16.33\%  & \underline{15.45\%}  & 15.73\%  & 16.20\%  & \textbf{15.03\%}  & 15.83\%   \\ 
    & RMSE &    32.87 & 31.24 & 31.59  & 30.85  & 31.12  & 31.29  & \textbf{30.73}  & \underline{30.77}   \\ 
    \midrule
    \multirow{3}{*}{PEMS-08} & MAE &    17.84 & 16.17 & 15.60  & 15.08  & 15.36  & 15.75  & \underline{14.91}  & \textbf{14.84}   \\ 
     & MAPE &   12.14\% & \textbf{10.59\%} & \underline{11.37\%}  & \underline{11.37\%}  & 12.67\%  & 14.23\%  & 12.53\%  & 12.40\%   \\ 
     & RMSE &   26.97 & 24.66 & 24.60  & 23.97  & 24.19  & 24.74  & \underline{23.72}  & \textbf{23.70}   \\ 
     \midrule
     \multirow{3}{*}{PEMS-BAY} & MAE &  2.01 & 1.71 & 1.60  & \textbf{1.58}  & 1.60  & 1.62  & 1.60  & \underline{1.59}   \\ 
     & MAPE &   4.65\% & 3.90\% & 3.60\%  & \textbf{3.53\%}  & \underline{3.57\%}   & 3.70\%  & \textbf{3.53\%}  & \underline{3.57\%}   \\ 
      & RMSE &   3.97 & \textbf{3.52} & 3.72  & \underline{3.67}  & 3.73  & 3.73  & 3.69  & 3.71   \\ 
    \midrule
    \multirow{3}{*}{METR-LA} & MAE & 3.43 & 2.97 & 2.89  & 2.90  & \underline{2.89}  & 2.99  & \textbf{2.88}  & 2.93   \\ 
     & MAPE &   10.02\% & 8.25\% & \textbf{8.13\%}  & \textbf{8.13\%}  & \underline{8.17\%}  & 8.40\%  & \textbf{8.13\%}  & 8.23\%   \\ 
     & RMSE &   6.61 & \textbf{5.78} & \underline{6.10}  & 6.12  & 6.12  & 6.26  & 6.11  & 6.21   \\ 

     \midrule
     \multicolumn{2}{c|}{Count} & \multicolumn{2}{c|}{5} & \multicolumn{4}{c|}{4} & \multicolumn{2}{c}{9}\\

    \bottomrule
  \end{tabular}
  }
    \caption{The spatial-temporal series forecasting results compared with popular NAS or Non-NAS baselines. The best performance is bold and the second best is underlined. Our method, ISTS, achieves the best performance in 9 of the fixed metrics and is highly competitive in the others}
      \label{Traffic-table}
\end{table*}
\section{Experiments}
We conduct extensive experiments on real-world datasets to evaluate the effectiveness of ISTS. More detailed settings and additional results are presented in~\ref{app-datasets} to~\ref{sec-case-study}. 

\subsection{Settings}
\subsubsection{Data Preparation}
\label{ch-data}

We conduct experiments on five real-world spatial-temporal sequence forecasting datasets, including PEMS-03/04/08~\cite{PEMS03}, PEMS-BAY~\cite{PEMSBAY} and METR-LA~\cite{METRLA}. The sample frequency is every 5 minutes, and we consider the input and output length as 12 sample points, which means the predicted duration is one hour. The basic information is summarized as Table~\ref{table:data-description}. Notably, we reduce the training subset for PEMS-BAY when architecture searching as the huge amount of data will cause memory exceed with gradient-based searching methods. To ensure the equality, we adopt the same partition for all methods in our experiments. More details of the dataset are presented in ~\ref{app-datasets}.

\begin{table}[t]

  \centering

  \begin{tabular}{cccccccccccc}
    \toprule
    \multirow{2}{*}{Dataset} & \multirow{2}{*}{Nodes} & \multirow{2}{*}{Steps} & \multicolumn{2}{c}{Partition (Train/Valid/Test)}  \\
    \cmidrule{4-5}
    &&& Search & Evaluation\\
    \midrule
    PEMS-03 & 358 & 26,139 & 7/1/2 & 7/1/2 \\
    PEMS-04 & 307 & 16,923 & 7/1/2 & 7/1/2 \\
    PEMS-08 & 170 & 17,787 & 7/1/2 & 7/1/2 \\
    PEMS-BAY & 325 & 52,047 & 4/4/2 & 7/1/2\\
    METR-LA & 207 & 34,203 & 7/1/2 & 7/1/2\\
    \bottomrule
  \end{tabular}
    \caption{Summary of basic information for spatial-temporal datasets. Nodes indicates the number of nodes in the spatial graph, and steps indicates the number of temporal steps in the whole data sequence.}
      \label{table:data-description}
\end{table}

\subsubsection{Baselines}
\label{ch-baseline}

We compare our method with several NAS methods and non-NAS methods as following. 
We adopted two \textbf{Non-NAS} spatial-temporal modeling methods as baselines evaluating with six datasets:
\begin{itemize}
    \item STSGCN\footnote[1]{STSGCN: https://github.com/j1o2h3n/STSGCN/tree/main}~\cite{STSGCN} uses a spatial-temporal synchronous module to extract spatial-temporal information, which is effective in capturing the complex localized spatial-temporal correlations.
    \item GraphWaveNet\footnote[2]{GraphWaveNet: https://github.com/nnzhan/Graph-WaveNet}~\cite{GraphWaveNet}  develops a novel adaptive dependency matrix and learn it through node embedding, which enables the model to precisely capture the hidden spatial dependency in the data.
\end{itemize}
We implement four \textbf{NAS methods} with the search space proposed in section~\ref{ch-search-space}.
\begin{itemize}
    \item DARTS\footnote[3]{DARTS: https://github.com/quark0/darts}~\cite{DARTS} is the pioneer in using gradient-based neural architecture search method and well-known framework. 
    \item PC-DARTS\footnote[4]{PC-DARTS: https://github.com/yuhuixu1993/PC-DARTS}~\cite{PC-darts} reduces the redundancy in exploring the network space by sampling. 
    \item SDARTS\footnote[5]{SDARTS: https://github.com/xiangning-chen/SmoothDARTS}~\cite{SDarts} smooths the loss landscape to improve the generalization ability. 
    \item DARTS-PT\footnote[6]{DARTS-PT: https://github.com/ruocwang/darts-pt}~\cite{PT} appends a perturbation-based selection to alleviate robustness issues. 
\end{itemize}

\subsubsection{Large Language Model} Our framework is designed to easily integrate newly released open-source large language models. In our experiments, we conducted experiments mainly based on LLaMA-3 with 8 billion parameters. The parameters of LLM are frozen and the pipeline is inference-only to make the search process more efficient. To assess the impact of different language models on our framework, we conduct comparative experiments based on Qwen2 (with 7 billion parameters) and Phi3-mini (with 3.8 billion parameters). 

All the experiments, including LLM's inferring, architecture simulating and the pipeline controlling, are conducted on an NVIDIA V100 GPU.

\subsubsection{Search Space Settings} The architecture search is based on a hybrid assembly way mentioned in Section~\ref{ch-search-space} with $4$ hidden states and $6$ cells, and each cell has 3 operator options (STP, STT and TTS). The total search rounds in our methods is $15$, that is to find well-performed architectures within $15$ rounds. From our experiments, $15$ rounds is enough for our methods to efficiently find competitive results for all of the downstream tasks. The exploring ratio is set to 0.6, which is suitable to explore enough architecture to reduce the possibility to be trapped in local optimal. The discussion of the exploration rate and search round can be found in Section~\ref{sec-hyper-parameter}.

\begin{table*}[htbp!]

  \centering
\resizebox{\linewidth}{!}{
  \begin{tabular}{cccccccccccc}
    \toprule
    \multirow{2}{*}{Method} & \multirow{2}{*}{DARTS} & \multicolumn{2}{c}{DARTS-PT} & \multirow{2}{*}{SDARTS} & \multirow{2}{*}{PC-DARTS} & \multicolumn{3}{c}{ISTS-COT(ours)} & \multicolumn{3}{c}{ISTS-TOT(ours)} \\
    \cmidrule(r){3-4}
    \cmidrule(r){7-9}
    \cmidrule(r){10-12}
    & & Search & Proj. & & & Sim. & Search & Total & Sim. & Search & Total \\
    \midrule
    PEMS-03 & 3029  & 3315  & 2974  & 3263  & 3348  & 208  & 17  & 225  & 207  & 84  & 291\\

    PEMS-04 & 1903  & 1922  & 1704  & 1954  & 1912  & 112  & 16  & 128  & 111  & 80  & 191\\

    PEMS-08 & 887  & 1177  & 1072  & 911  & 955  & 57  & 12  & 68  & 58  & 89  & 147\\

    PEMS-BAY$^*$ & 3865  & 3852 & 3462 & 3931  & 3914  & 229  & 17  & 245  & 232  & 79  & 311 \\

    METR-LA & 4018  & 4033 & 3670 & 4022  & 4132  & 181  & 16  & 196  & 180  & 78  & 258   \\

    \bottomrule
  \end{tabular}
  }
    \caption{The efficiency comparison of NAS methods, evaluating processing seconds per epoch. DARTS-PT has two stages, namely search and projection (Proj.). The projection process starts after all the search process ends, so the speed could not be added. ISTS also has two stages including LLM instructed architecture search and one-step tuning simulation (Sim.). ($^*$We evaluate PEMS-BAY with only 40\% data used for search process for all methods in case of the memory exceed for gradient-based methods.)}
      \label{table:efficiency}
\end{table*}

\begin{figure}
  
\centering
\includegraphics[width=\linewidth]{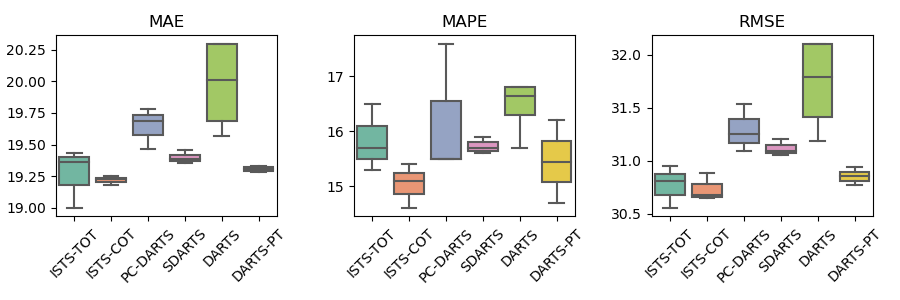}
   \caption{The performance comparison on the PEMS-04 dataset indicating that our method consistently achieves high  performance in MAE and RMSE and is competitive in MAPE. 
   }
   \label{box-fig}

\end{figure}

\subsection{Performance of Searched Architecture} 
\label{ch-res-main}

To evaluate the ability of our searching method with LLM, we make a comparison on the performance of searched architectures on five datasets. Our method show competitive performance compared to six popular NAS and Non-NAS methods, achieving the best performance in 9 of the fixed metrics on 5 datasets.  

Table~\ref{Traffic-table} summaries of the results evaluating on 5 datasets, comparing the mean absolute error (MAE), mean absolute percentage error (MAPE), and root mean square error (RMSE), and the definition of metrics can be found in~\ref{app-metrics}. We report the average performance with 3 repeated experiments. 

Figure \ref{box-fig} shows the performance distribution of different NAS methods on dataset PEMS-04. Our method exhibits stable and advanced performance compared to SOTA baselines. Plots for other datasets can be found in~\ref{app-more-results}.

\subsection{Searching Efficiency}

We evaluate the efficiency with process speed in Table~\ref{table:efficiency}. Traditional NAS methods is time-consuming due to the policy learning process. However, our method, by utilizing LLM's internal knowledge and comprehensive ability for decision-making, avoids the time-consuming training process and can achieve competitive performance with inference-only LLMs. As a result, the efficiency of our method is significantly higher than traditional NAS methods. 

Specifically, policy learning with DARTS method in PEMS-08 consumes 887 seconds per epoch to learn the decision-making strategy in architecture searching, while ISTS-COT consumes only totally 68 seconds, including about 12 seconds for LLM inferring for architecture search and 57 seconds for architecture simulation, which reduce about 92\% of the processing time.   
\begin{table}[t]

  \centering
    \resizebox{\linewidth}{!}{
  \begin{tabular}{ccccccccccc}
    \toprule
    \multicolumn{2}{c}{Ablation}  & \makecell{ISTS} & $\neg$stra & $\neg$mem \\
    \midrule
    \multirow{3}{*}{PEMS-03} & MAE & 15.41 & 15.48 & 15.52\\
    & MAPE & 17.57\% & 17.77\% & 18.47\% \\
    & RMSE & 26.23 & 26.49 & 26.57\\
    \midrule
    \multirow{3}{*}{PEMS-04} & MAE & 19.22 & 19.28 & 19.37\\
    & MAPE & 15.03\% & 15.23\% & 15.83\%\\
    & RMSE & 30.73 & 30.85 & 30.88\\
    \midrule
    \multirow{3}{*}{PEMS-08} & MAE & 14.91 & 15.61 & 15.55\\
    & MAPE & 12.53\% & 12.85\% & 13.20\%\\
    & RMSE & 23.72 & 24.52 & 24.47 \\

    \bottomrule
  \end{tabular}
  }
    \caption{Ablation study. Removing the two-stage strategy and memory accumulation framework make the performance decline. }
      \label{Ablantion-table}
\end{table}

\subsection{Ablation Study }
\label{res-prompts}

In Table~\ref{Ablantion-table}, we assessed the method without certain components.  We conducted the ablation experiments to remove the two-stage strategy process and memory accumulation framework. 

Without two-stage strategy ($\neg$stra), the LLM optimize the performance from the first round without encouraging it to explore more promising spaces, which makes it easy to be trapped with local optima. 

Removing the memory accumulation framework ($\neg$mem), the ISTS must make decisions only by its internal knowledge without cumulated knowledge, and its ability is highly limited to LLM's comprehension ability, which leads to unstable performance.

We further analyzed the behaviors of the LLM with its responses in~\ref{sec-case-study}. And we found the strategies and memories highly influence LLM's decisions in architecture designing.


\begin{table}

  \centering
    \resizebox{\linewidth}{!}{
  \begin{tabular}{ccccccccccc}
    \toprule
    \multicolumn{2}{c}{\multirow{2}{*}{LLM}}  & LLaMA3 & Qwen2 & Phi3-mini  \\
    & & (8B) & (7B) & (3.8B) \\
    \midrule
    \multirow{3}{*}{PEMS-03} & MAE & 15.41 & 15.43 & 15.30 \\
    & MAPE & 17.57\% & 18.00\% & 18.43\%\\
    & RMSE & 26.23 & 26.18 & 26.09 \\
    \midrule
    \multirow{3}{*}{PEMS-04} & MAE & 19.22 & 19.26 & 19.20 \\
    & MAPE & 15.03\% & 15.47\% & 15.43\%\\
    & RMSE & 30.73 & 30.77 & 30.78\\
    \midrule
    \multirow{3}{*}{PEMS-08} & MAE & 14.91 & 15.18 & 15.20\\
    & MAPE & 12.53\% & 12.77\% & 12.97\%\\
    & RMSE & 23.72 &  24.05 & 24.08\\

    \bottomrule
  \end{tabular}
  }
    \caption{Results of the sensitivity of LLMs. Our method achieve stable performance across different LLMs.}
    \label{LLM-Sensitivity-table}
\end{table}

\subsection{Sensitivity of LLM}
To evaluate the sensitivity of LLM in our framework, we tested our method on 3 widely known open-source large language models, including LLaMA3 (LLaMA-3-8B-instruct), Qwen-2 (Qwen-2-7B-instruct), Phi-3 (Phi-3-mini-4k-instruct). 

The results are listed in Table \ref{LLM-Sensitivity-table}. Though the LLM's ability slightly affects the performance, the results show that our framework performs consistently well across different LLMs.  
\begin{figure}[t]
  
    \centering
    \begin{minipage}{\linewidth}
        \centering

            \includegraphics[width=\linewidth]{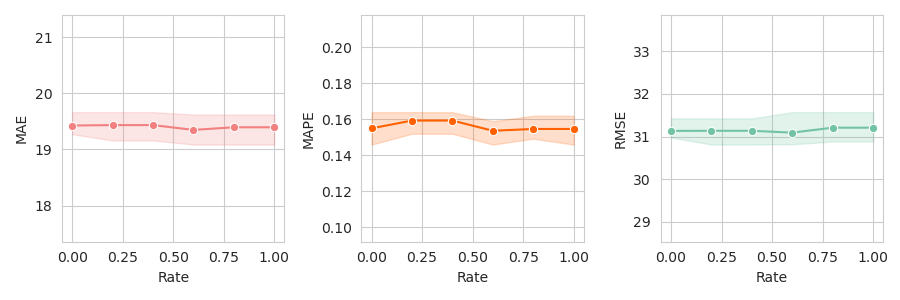}
            \caption{Evaluation of the exploration rates. }
            \label{fig-rate}
    \end{minipage}
        \begin{minipage}{\linewidth}
        \centering

        \includegraphics[width=\linewidth]{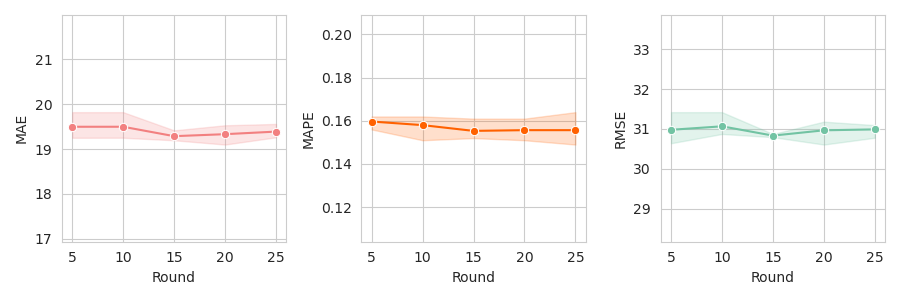}
        \caption{Evaluation of the searching rounds. }
        \label{fig-round}
    \end{minipage}

\end{figure}
\subsection{Hyper Parameter Exploration}
\label{sec-hyper-parameter}

We evaluates the performance with different exploration rates and searching rounds with dataset PEMS-04. 

Exploration rates constrains the number of rounds in exploration stages. We found setting exploration rates as 0.6, representing that using the first 60\% rounds to explore more search space rather than optimize, makes our model achieves best performance, which balance the exploration and optimization process. 

We evaluates the performance across different searching rounds in Figure~\ref{fig-round}. We found that when setting search rounds as 15, our model achieves best performance. Our method can achieve competitive performance in architecture searching with a few rounds, while learning-base NAS methods usually needs more than 30 rounds for policy learning and the number of learning rounds significantly affects the performance.

\begin{figure}[htbp!]
  
\centering
   \includegraphics[width=\linewidth]{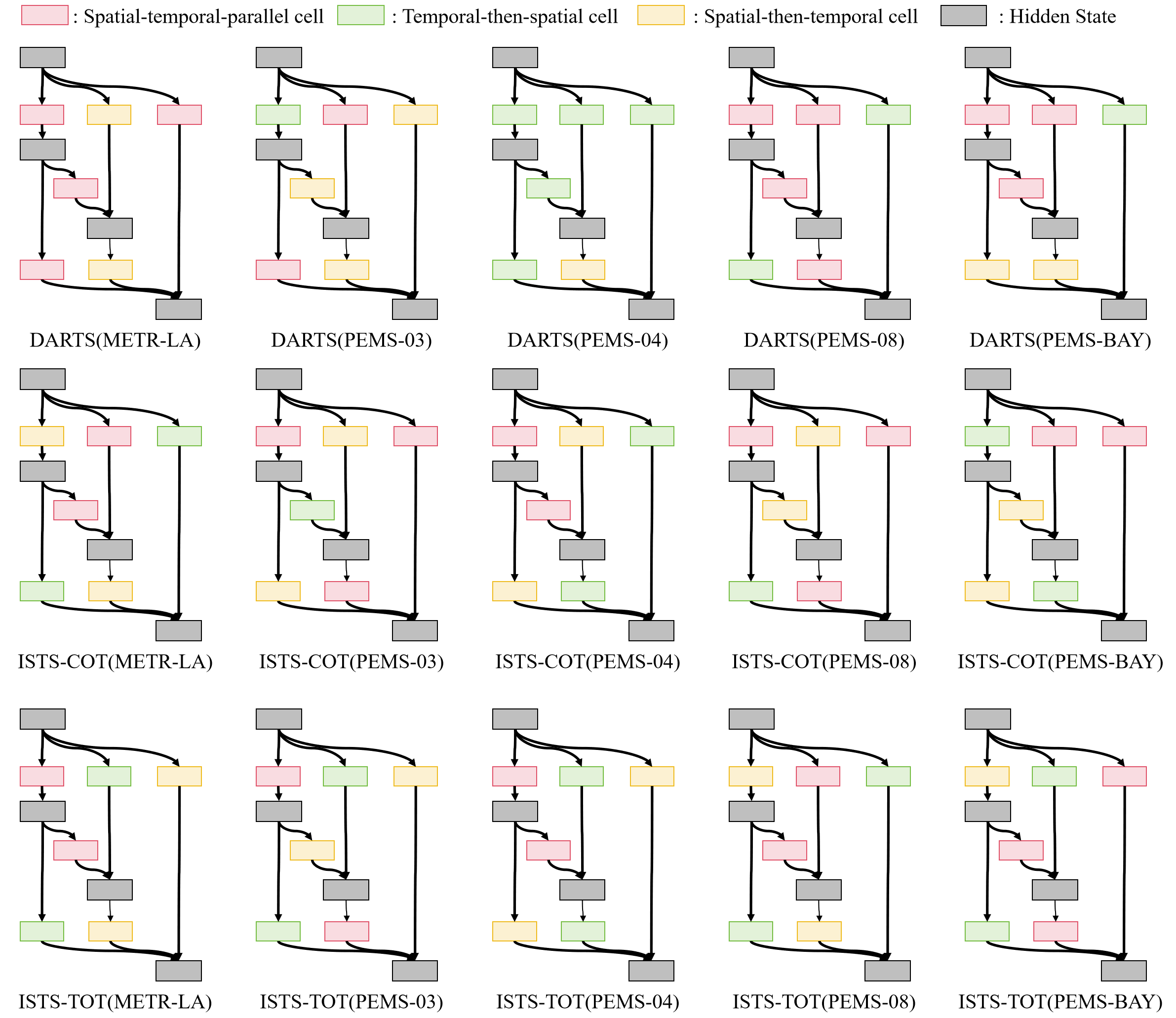}
   \caption{Visualization of the architectures discovered by DARTS and our method. }
   \label{archs}

\end{figure}

\subsection{Visualization}

We visualize parts of the architecture searched by DARTS and our method on the five dataset in Figure \ref{archs}.

Specifically, for dataset PEMS-04, DARTS framework build an architecture with five temporal-then-spatial cell along with a spatial-then-temporal cell. We can indicate that DARTS prefer to find architectures with cells' types relatively homogeneous, which is conducive for policy learning based on gradient optimization. However, LLMs are not subject to such constraints, so the architecture found by LLM can be more various which makes our method more likely to quickly find the global optimal solution. This can explain why our method outperforms traditional NAS methods on this dataset.

\section{Conclusion}
To solve the challenge of relying on empirical knowledge in building and tuning architecture search frameworks, we propose a multi-level enhanced LLM-based neural architecture search method for spatial-temporal sequence forecasting. At the instance-level, we decompose the architecture search task with powerful prompts and guide LLM to find promising candidates sensibly. At the instance-level, we introduce a quick simulation and evaluation framework to provide feedback and experience accumulation to enhance effectiveness and ensure efficiency. At the task-level, we present the two-stage searching mechanism to reduce the possibility of trapped with local optima. Experiment results show that ISTS achieves better or competitive performance compared to state-of-the-art methods with superior efficiency. We also evaluated the influence of the hyper-parameters for its determination. We discussed our method's sensitivity of LLM and our method show stability though LLM's ability has difference. We further visualized the architectures found by different methods to explore the ability of avoid falling into local optima.

\section{Acknowledgement}
This work was supported by the grants from the Natural Science Foundation of China (62202029), and Young Elite Scientists Sponsorship Program by CAST (No. 2023QNRC001). Thanks for the computing infrastructure provided by Beijing Advanced Innovation Center for Big Data and Brain Computing. Jianxin Li is the corresponding author.

\clearpage

\appendix

\section{Datasets}
\label{app-datasets}

In our study, we mainly focus on the architecture searching for spatial-temporal sequence forecasting (STSF). The datasets involved in our experiments are listed as follows.

\textbf{METR-LA~\cite{METRLA}} The traffic speed dataset consists of vehicle speed collected with 207 detectors located the highways of Los Angeles County, USA. The dataset contains 5-minute grained data, ranging from March 1, 2012 to June 30, 2012. The distances between detectors are provided as spatial information.

\textbf{PEMS-BAY/03/04/08~\cite{PEMSBAY}} The traffic flow datasets originate from the Performance Evaluation System (PeMS) of the California Transportation Agencies (CalTrans). All data share a time interval of 5 minutes. The distances based on real road networks between sensors are provided as spatial information. 

\section{Metrics}
\label{app-metrics}

We evaluate the performance of searched architectures with 3 metrics, including MAE, MAPE and RMSE, following the previous works \cite{STSGCN, GraphWaveNet}.

\textbf{ Mean Absolute Error (MAE).} The MAE metrics is shown as follow:
\begin{equation}
    \text{MAE}(\hat{Y}, Y) = \frac{1}{M} \sum_{i=1}^{M}|\hat{Y}_i - Y_i|,
\end{equation}

where $\hat{Y}, Y$ are overall prediction and real value respectively; $\hat{Y}_i , Y_i$ are predicted and real value of one data window respectively; $M$ is the count of data windows.

\textbf{ Mean Absolute Percentage Error (MAPE).} The MAPE metrics is shown as follow:
\begin{equation}
    \text{MAPE}(\hat{Y}, Y) = \frac{100\%}{M} \sum_{i=1}^{M}\frac{|\hat{Y}_i - Y_i|}{Y_i},
\end{equation}

where $\hat{Y}, Y$ are defined as above.
MAPE is a percentage variation of MAE. MAPE divides the variation with the scale of original data, which is less sensitive to anomaly values. 

\textbf{ Root Mean Square Error (RMSE).} The RMSE metrics is shown as follow:
\begin{equation}
    \text{RMSE}(\hat{Y}, Y) = \sqrt{\frac{1}{M} \sum_{i=1}^{M}(\hat{Y}_i - Y_i)^2},
\end{equation}

where $\hat{Y}, Y$ are defined as above.
RMSE is the square root of the MSE, whose scale is the same as the original data values. 

\section{Hyper Parameters Setting}

More hyper-parameters of the spatial-temporal cells are listed in Table~\ref{table-hyper-parameters}. The three kinds of cell share these hyper parameters in a certain downstream task. We explain these hyper parameters as follow: 

\begin{itemize}
    \item \textbf{Batch Size (B.S.).} The number of instances each batch.
    \item \textbf{Hidden Size (H.S.).} The dimension of hidden stages.
    \item \textbf{Attention Dimension (Attn. Dim.).} The dimension of attention mechanism. Attention is mainly use to capture temporal dependencies in spatial-temporal cells.
    \item \textbf{Attention Head (Attn. Head).} The number of heads use in attention mechanism. 
    \item \textbf{FFN Dimension (FFN Dim.).} The dimension of feed-forward networks. 
    \item \textbf{Graph order (Gra. Ord.).} The hyper parameter in high-order spatial operator. 
    \item \textbf{Dropout (Drop.).} The dropout rate of architecture.

\end{itemize}

\begin{table}[hbtp!]

  \centering
\resizebox{\linewidth}{!}{
  \begin{tabular}{ccccccccccc}
    \toprule
    Dataset & \makecell{B.S.} & \makecell{H.S.} & \makecell{Attn.\\Dim.} & \makecell{Attn.\\Head} & \makecell{FFN\\Dim.} & \makecell{Gra.\\Ord.} & Drop. \\
    \midrule
    PEMS-03 & 16 & 32 & 32 & 8 & 64 & 2 & 0.1 \\
    PEMS-04 & 16 & 64 & 64 & 8 & 64 & 2 & 0.1 \\
    PEMS-08 & 32 & 64 & 64 & 8 & 64 & 2 & 0.1 \\
    PEMS-BAY & 16 & 64 & 64 & 8 & 256 & 2 & 0.1 \\
    METR-LA & 16 & 64 & 64 & 8 & 256 & 2 & 0.1 \\
    \bottomrule
  \end{tabular}
  }
    \caption{Hyper-parameters for the spatial-temporal cells. }
      \label{table-hyper-parameters}
\end{table}

\section{Details of the Prompts}
\label{app-prompts}

The input of LLM includes the background information and instructive prompts.

The \textbf{backgrounds} part provides information including datasets and metrics to inspire LLM to recall relevant internal knowledge. Besides, we describe the cells and the architecture in this part, to inform LLM with the search unit and the search space. The background information, shown as Table~\ref{table-background} is shared in each round, no matter which stage or thought type.

The \textbf{instruction} part introduces what should LLM do, which is the shown in Table~\ref{table-instructioins}. The instructions are different depends on current stages and thought type. For exploration stage, the instructions encourage the LLM to explore new combination that is not existed in historical samples to potentially achieve better performance. For optimization stage, the prompts set the target for LLM, which is to find the best combination according to previous tried samples to make metrics as low as possible. The COT type decomposition only concentrates on the generation process of the whole architecture with several thought steps. While the TOT type decomposition further consider the choosing of each cell, so it has generation steps and evaluation steps to gradually generate the whole architecture cell by cell. Besides, we define the response format for LLM. Only the response strictly compiled to the defined response format will be treated as a legal response. The inter-wares will act according to the instructions recognized in the response.

\begin{table*}

  \centering

  \begin{tabularx}{\textwidth}{l|X}
    \toprule
    \multirow{7}{*}{Backgrounds}  & Please select appropriate modules for the following deep learning task. \\
    
    &  \textbf{Background description:} \\
    &  \textbf{The dataset:} This is a traffic forecasting dataset, consisting of hundreds of sensors monitoring the traffic indices around the city. The goal is to predict future traffic indices according to history indices for each sensor. \\
    &  \textbf{The options:} The deep learning task should properly capture spatial and temporal information with a certain combination of several layers, and each layer can choose one of the three kinds of modules, namely spatial-then-temporal, temporal-then-spatial and spatial-temporal-parallely. \\
    &  \textbf{The architecture:} The whole deep learning architecture is a directed acyclic graph with four nodes and six layers, and the preceding nodes are connected to each subsequent nodes with a layer. e.g. Node 1 connect Nodes 2 with Layer 1, Node 1 connect Nodes 3 with Layer 2, ... \\
    &  \textbf{The metrics:} Three metrics to evaluate the results predicted by the model, including MAE (Mean Absolute Error), MAPE (Mean Absolute Percentage Error), and RMSE (Root Mean Square Error) \\
    &   Here are some historical samples obtained in previous rounds to guide you to select more suitable combination of modules (sorted by MAE):  \textcolor{blue}{\{samples\}} \\
        \bottomrule
  \end{tabularx}
    \caption{Prompts for backgrounds, sharing in any stages and steps. }
      \label{table-background}
\end{table*}
\begin{table*}

  \centering

  \begin{tabularx}{\textwidth}{c|c|c|X}
    \toprule
    Stage & \makecell{Thought \\Type} & \makecell{Instruction \\Type} & {Prompts} \\
    \midrule
    \multirow{20}{*}{\makecell{Exploration\\ Stage \\($S_1$)}} & \multirow{4}{*}{\makecell{COT \\ ($P_C$)}} & \multirow{4}{*}{\makecell{Generation\\($P_{CG}$)}} & \textbf{Your task:} \\
    & & &    First, analyze the background description of this task.  \\
    & & &   Then, observe the samples from previous rounds, consider the applicability of different modules in this task. \\
    & & &   Next, considering these factors comprehensively, for the six layers, try to design a new combination that is not existed in historical samples to potentially achieve better performance and explain your choice. You have \textcolor{blue}{\{total\_epoch\}} rounds to try, and this is the \textcolor{blue}{\{current\_epoch + 1\}} round.  \\
    & & & Provide no additional text in response, Format output in JSON as \{\{ "Combination of modules": \{\{"Layer\_1": "choice for layer\_1", "Layer\_2": "choice for layer\_2", "Layer\_3": "choice for layer\_3", "Layer\_4": "choice for layer\_4", "Layer\_5": "choice for layer\_5", "Layer\_6": "choice for layer\_6"\}\}, "Explanation": "explain your choice"\}\}
      \\
    \cmidrule{2-4}
    & \multirow{12}{*}{\makecell{TOT\\($P_T$)}} & \multirow{6}{*}{\makecell{Generation\\($P_{TG}$)}} &  You have chosen \textcolor{blue}{\{current\_layer\_num\}} layers, they are: \textcolor{blue}{\{current\_layers\}}. \\
    & & & \textbf{Your task:} \\
    & & &  First, analyze the background description of this task.  \\
    & & &    Then, observe the historical samples and the layers you have chosen.\\ 
    & & &  Next, considering these factors comprehensively, try to choose the next one layer based on your current chosen layers, that should not be too similar to the history samples to potentially achieve better performance, and explain the reason. \\
    & & & Provide no additional text in response, Format output in JSON as \{\{"New layer": "Your choice for new layer", "Explanation": "explain your choice"\}\}\\
    \cmidrule{3-4}
    & & \multirow{6}{*}{\makecell{Evaluation\\($P_{TE}$)}} &  You have chosen \textcolor{blue}{\{current\_layer\_num\}} layers, they are: \textcolor{blue}{\{current\_layers\}}. \\
    & & & \textbf{Your task:} \\ 
    & & &    First, analyze the background description of this task.  \\
    & & &    Then, observe the historical samples and the layers you have chosen.\\ 
    & & &    Next, judge if it is possible that the layers you have chosen will lead into a better result and explain the reason.\\
    & & & Provide no additional text in response, Format output in JSON as \{\{"Judgment": "possible or impossible", "Explanation": "explain your judgment"\}\}\\
    \midrule
    \multirow{5}{*}{\makecell{Optimization \\Stage \\$S_2$}} &\multirow{5}{*}{\makecell{COT/TOT\\($P_C$/$P_T$)}} & \multirow{5}{*}{\makecell{Generation\\($P_{CG}$/$P_{TG}$)}}& \textbf{Your task:} \\
    & & &    First, analyze the background description of this task.  \\
    & & &   Then, observe the samples from previous rounds, consider the applicability of different modules in this task. \\
    & & &   Next, considering these factors comprehensively, for the six layers, try to find the best combination according to previous tried samples to make MSE, MAE and RMSE lower and explain your choice. You have \textcolor{blue}{\{total\_epoch\}} rounds to try, and this is the \textcolor{blue}{\{current\_epoch + 1\}} round. 
     \\
    & & & Provide no additional text in response, Format output in JSON as \{\{ "Combination of modules": \{\{"Layer\_1": "choice for layer\_1", "Layer\_2": "choice for layer\_2", "Layer\_3": "choice for layer\_3", "Layer\_4": "choice for layer\_4", "Layer\_5": "choice for layer\_5", "Layer\_6": "choice for layer\_6"\}\}, "Explanation": "explain your choice"\}\}
      \\
    \bottomrule
  \end{tabularx}
    \caption{Prompts for instructions, different across stages and steps. }
      \label{table-instructioins}
\end{table*}

\section{More Results}
\label{app-more-results}

\begin{figure}
  
\centering
    \begin{minipage}{\linewidth}
    \includegraphics[width=\linewidth]{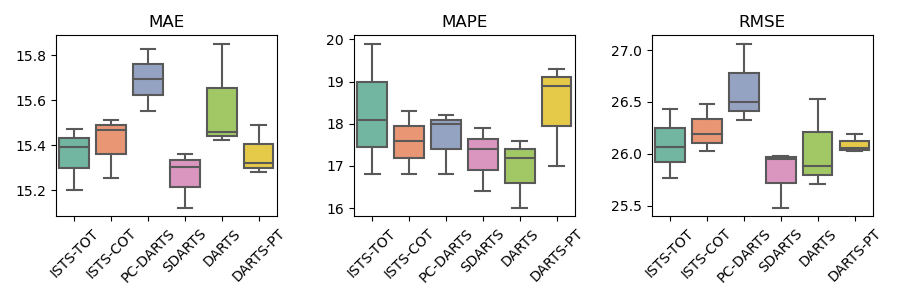}
   \caption{The performance comparison on the PEMS-03 dataset. 
   }
   \medskip
   \label{box-fig-03}
   \end{minipage}
   \begin{minipage}{\linewidth}
   \includegraphics[width=\linewidth]{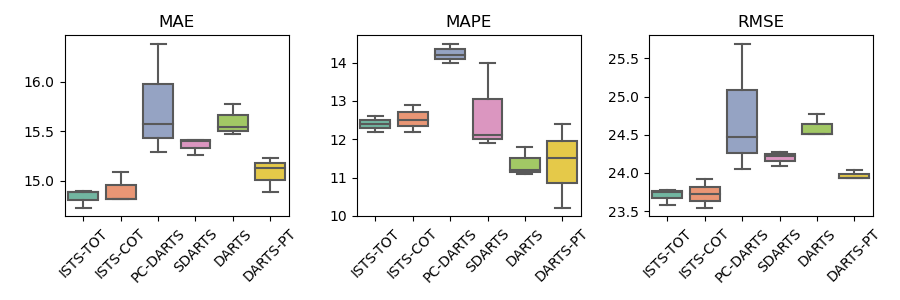}
   \caption{The performance comparison on the PEMS-08 dataset. 
   }
   \label{box-fig-08}
   \end{minipage}
      \begin{minipage}{\linewidth}
      \includegraphics[width=\linewidth]{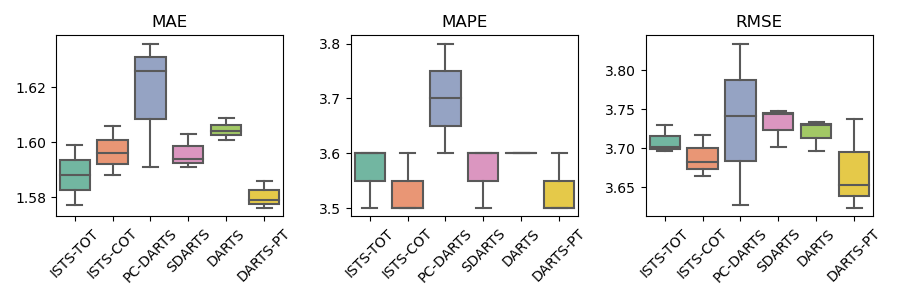}
   \caption{The performance comparison on the PEMS-BAY dataset. 
   }
   \medskip
   \label{box-fig-BAY}
   \end{minipage}
      \begin{minipage}{\linewidth}
      \includegraphics[width=\linewidth]{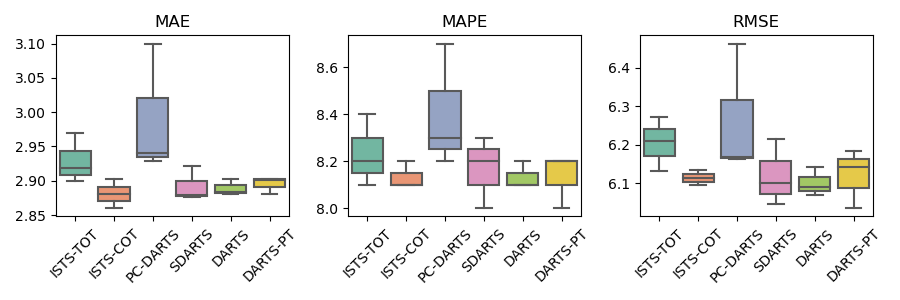}
   \caption{The performance comparison on the METR-LA dataset. 
   }
   \label{box-fig-LA}
   \end{minipage}

\end{figure}

Figure~\ref{box-fig-03}, \ref{box-fig-08} ,\ref{box-fig-BAY}, \ref{box-fig-LA} shows the performance of PEMS-03/08/BAY and METR-LA, the same format with Figure~\ref{box-fig} (for PEMS-04). We use the box plots to visually show the distribution of the results and comparing different methods. The figures indicate the stability of our methods in most of the metrics compared to traditional NAS methods.

\section{Examples of LLM's response}
\label{sec-case-study}

We select several responses of LLM, including its search decisions and the self explanation of its behaviors. We can infer that, the memories and strategies influence LLM's behaviors and instructing its to make reasonable decisions.   

Table~\ref{table-COT-example} contains examples of LLM's response with COT type templates. LLM responses an entire architecture each round and explain the reason of choosing each layer. The examples in Table~\ref{table-COT-example} is chosen from the round 1 and round 12 of a searching instance guided by LLaMA-3 when considering dataset PEMS-04. For the round 1, LLM makes decisions mainly based on background description as the samples accumulated from previous rounds are limited. For the round 12, LLM not only considers background information but also seek inspiration from existing samples from previous rounds.

Table~\ref{table-TOT-examples} contains examples of LLM's response with TOT type templates. LLM responses a cell of the architecture each step, and a round consists of 6 steps to form a completed architecture. The examples in Table~\ref{table-TOT-examples} is chosen from an exploration round of a searching instance guided by LLaMA-3 when considering dataset PEMS-04. For step 1 of this round, LLM mainly considers to decide whether to explore newly architectures which is never existed. For step 2 of this round, as the decision of whether to explore direction has been made in previous steps, LLM only need to make further analysis with existing information and current choosing layer.

\begin{table*}

  \centering
  \begin{tabularx}{\textwidth}{c|X}
    \toprule
    Round & \makecell{Response} \\
    \midrule
    
    \multirow{16}{*}{1}
    & LLM response: \{ \\
    &   \quad "Combination of modules": \{ \\
    &   \qquad "Layer\_1": "spatial-temporal-parallel", \\
    &   \qquad "Layer\_2": "temporal-then-spatial", \\
    &   \qquad "Layer\_3": "spatial-then-temporal", \\
    &   \qquad "Layer\_4": "spatial-temporal-parallel", \\
    &   \qquad "Layer\_5": "temporal-then-spatial", \\
    &   \qquad "Layer\_6": "spatial-then-temporal" \\
    &   \quad \}, \\
    &   \quad "Explanation": "\textcolor{red}{Based on the background description}, the task requires capturing both spatial and temporal information. The combination of spatial-temporal-parallel in Layer\_1 allows the model to process both spatial and temporal information simultaneously. The subsequent layers, temporal-then-spatial and spatial-then-temporal, help to refine the predictions by considering the temporal and spatial relationships between the sensors. This combination is not present in the historical samples, and it may provide a better balance between spatial and temporal information." \\
    & \} \\
    \midrule

    \multirow{16}{*}{12} 
    & LLM response: \{ \\
    &   \quad "Combination of modules": \{ \\
    &   \qquad "Layer\_1": "spatial-temporal-parallel", \\
    &   \qquad "Layer\_2": "spatial-then-temporal", \\
    &   \qquad"Layer\_3": "spatial-temporal-parallel", \\
    &   \qquad "Layer\_4": "temporal-then-spatial", \\
    &   \qquad "Layer\_5": "spatial-then-temporal", \\
    &   \qquad "Layer\_6": "spatial-temporal-parallel" \\
    &   \quad\}, \\
    &   \quad "Explanation": "\textcolor{red}{Based on the analysis of the background description and the samples from previous rounds}, I choose the combination of modules that has the lowest MAE and RMSE. The combination of spatial-temporal-parallel, spatial-then-temporal, and temporal-then-spatial modules seems to be effective in capturing both spatial and temporal information. The combination of spatial-temporal-parallel and spatial-then-temporal modules in the first three layers helps to capture the spatial information, while the combination of temporal-then-spatial modules in the last three layers helps to capture the temporal information. This combination has been tried in several previous rounds and has shown good performance." \\
    &   \} \\
    \bottomrule
  \end{tabularx}
    \caption{Examples of LLM's response with COT type templates.}
      \label{table-COT-example}
\end{table*}

\begin{table*}

  \centering
  \begin{tabularx}{\textwidth}{c|c|X}
    \toprule
    Step & \makecell{Response \\Type} & \makecell{Response} \\
    \midrule
    \multirow{16}{*}{1} & \multirow{8}{*}{Layer Choice} & 
    LLM response: \{ \\
    & & \quad "New layer": "spatial-then-temporal", \\
    & & \quad "Explanation": "\textcolor{red}{Based on the historical samples, it seems that the combination of spatial-then-temporal and temporal-then-spatial modules has been tried before.} To potentially achieve better behaviors, I choose a new combination that is different from the previous ones. Spatial-then-temporal module can capture the spatial information and then use temporal information to make predictions, which is a common and effective approach in traffic forecasting tasks." \\
    & & \} \\
    \cmidrule{2-3}
    &  \multirow{9}{*}{Evaluation} & 
    LLM eval: \{ \\
    & & \quad "Judgment": "possible", \\
    & & \quad "Explanation": "\textcolor{red}{Based on the historical samples, it seems that the combination of spatial-then-temporal and temporal-then-spatial modules has been tried before, and the results are not consistently improving.} The combination of spatial-temporal-parallel modules has also been tried, but with varying results. The chosen combination of \{'Layer\_1': 'spatial-then-temporal'\} is a new combination that has not been tried before. Although it's difficult to predict the exact outcome, it's possible that this combination could lead to a better result, especially since it's a new combination that hasn't been tried before." \\
    & & \} \\
    \midrule
    \multirow{16}{*}{2} & \multirow{8}{*}{Layer Choice} &
    LLM response: \{ \\
    & & \quad "New layer": "temporal-then-spatial", \\
    & & \quad "Explanation": "\textcolor{red}{Considering the background description, the task requires capturing both spatial and temporal information.} The chosen layer 'spatial-then-temporal' is a good starting point. To build upon this, I choose 'temporal-then-spatial' as the next layer. This is because 'temporal-then-spatial' can help to further process the temporal information and then apply spatial information, which can be beneficial for traffic forecasting. Additionally, this choice is not too similar to the history samples, as Round 2 and Round 3 have already tried 'spatial-then-temporal' and 'spatial-temporal-parallel' respectively, so 'temporal-then-spatial' can provide a different perspective." \\
    & & \} \\
    \cmidrule{2-3}
    &  \multirow{8}{*}{Evaluation} & 
    LLM eval: \{ \\
    & & \quad "Judgment": "possible", \\
    & & \quad "Explanation": "The combination of 'spatial-then-temporal' and 'temporal-then-spatial' layers can capture both spatial and temporal information. \textcolor{red}{The historical samples} show that the combination of these two types of layers has been tried before, and the MAE has been decreasing. Although the MAE has not decreased significantly in the last two rounds, it's possible that the combination of these two layers can still lead to a better result, especially considering that the MAE in Round 0 is relatively high. The reason is that the combination of 'spatial-then-temporal' and 'temporal-then-spatial' layers can effectively capture the spatial and temporal dependencies in the traffic forecasting dataset." \\
    & & \} \\

    \bottomrule
  \end{tabularx}
    \caption{Examples of LLM's TOT type templates.}
      \label{table-TOT-examples}
\end{table*}

\bibliographystyle{elsarticle-num} 

\end{document}